# Reactive Collision Avoidance using Evolutionary Neural Networks

Hesham M. Eraqi[1], Youssef Emad Eldin[2] and Mohamed N. Moustafa[1]

[1] Department of Computer Science and Engineering, The American University in Cairo, New Cairo 11835, Egypt

[2] Department of Computer and Systems Engineering, Ain Shams University, Cairo, Egypt

hesham.eraqi@gmail.com, youssef.emad.attia@gmail.com, m.moustafa@aucegypt.edu

Abstract: Collision avoidance systems can play a vital role in reducing the number of accidents and saving human lives. In this paper, we introduce and validate a novel method for vehicles reactive collision avoidance using evolutionary neural networks (ENN). A single front-facing rangefinder sensor is the only input required by our method. The training process and the proposed method analysis and validation are carried out using simulation. Extensive experiments are conducted to analyse the proposed method and evaluate its performance. Firstly, we experiment the ability to learn collision avoidance in a static free track. Secondly, we analyse the effect of the rangefinder sensor resolution on the learning process. Thirdly, we experiment the ability of a vehicle to individually and simultaneously learn collision avoidance. Finally, we test the generality of the proposed method. We used a more realistic and powerful simulation environment (CarMaker), a camera as an alternative input sensor, and lane keeping as an extra feature to learn. The results are encouraging; the proposed method successfully allows vehicles to learn collision avoidance in different scenarios that are unseen during training. It also generalizes well if any of the input sensor, the simulator, or the task to be learned is changed.

## 1 INTRODUCTION

The task of designing control software for a self-driving car is a complex task. The software should concurrently tolerate (model) infinite number of scenarios and special cases, and maintain and meet reasonable software complexity and resources constrains. Evolutionary algorithms can be a good alternate to abstraction from such control challenges (Sipper, 2006).

Collision avoidance is a feature that allows a vehicle to move without colliding with other vehicles. Vehicles can be cars, trains, ships, airplanes, Unmanned Aerial Vehicles (UAV), or various smart robots that have been generally applied in modern laboratories nowadays (Liu et al., 2013). In many applications, collision avoidance systems play a vital role in reducing the number of accidents and saving human lives. Reactive collision avoidance controls the motion of the vehicle directly based on the current sensor data to react to unforeseen changes in unknown and dynamic environments. The dynamic objects and the static environment do not cooperate with the ego-vehicle (vehicle that learns) to achieve collision avoidance.

Hence, reactive collision avoidance has a good performance in real-time (Fu et al., 2013).

We introduce a novel method for vehicles reactive collision avoidance using evolutionary neural networks (ENN). A single front-facing rangefinder sensor is the only input required by our method. The sensor provides the neural network with spatial proximity readings measured at multiple horizontal angles. The neural network learns how to control the vehicle steering wheel angle by directing the vehicle such that it does not collide with the dynamic environment. The neural network guides the vehicle around the environment and a genetic algorithm is used to pick and breed generations of more intelligent vehicles. The training process and the proposed method analysis and validation are carried out using simulation.

We conducted six experiments to validate the proposed method, analyse evaluate its performance. The results are encouraging; the proposed method successfully allows vehicles to learn collision avoidance in different scenarios that are unseen during training. The scenarios include a vehicle that learns how to safely navigate (without doing collision) through a free static track and to achieve collision avoidance among independent dynamic vehicles. Also, a group of randomly moving vehicles successfully learns how to achieve collision avoidance simultaneously. Also, our method is proven to generalize well, it successfully allows vehicle to also learn lane keeping, and using different simulation environment which is more realistic and powerful: CarMaker (CarMaker open test platform for virtual test driving website).

The disadvantage of traditional methods over our method are mainly: 1) they either depend on defined set of scenarios, which are not adapted to new conditions not programmed in the algorithm or 2) they rely on handcrafted features that do not well represent the real scenarios where the vehicle is deployed. This creates the need to new AI systems that learn from data, and in the same time automatically identify the best representations of this environmental data. Neural networks are well known for their ability to learn representations of the data.

## 2 RELATED WORK

(Shaffer et al.,1992) in "Combinations of Genetic Algorithms and Neural Networks: A Survey of Art" provided an overview of the literature of combining Neural Networks and genetic Algorithms drawing out the common themes and the emerging wisdom about what seems to work and what does not.

(Montana and L. Davis, 1989) in "Training feedforward neural networks using genetic algorithms" has explained that multilayered feedforward neural networks possess a number of properties which make them particularly suited to complex pattern classification problems and showed that Genetic Algorithms are well suited to the problem of training feedforward networks as they are good at exploring a large and complex space in an intelligent way to find values close to the global optimum.

(Durand et al, 1996) in "collision avoidance using neural networks learned by genetic algorithms" handled the collision avoidance problem between two aircrafts with reactive techniques using neural networks which was built by genetic algorithms.

(Togelius and Lucas, 2006) in "Evolving robust and specialized car racing skills" presented using evolutionary algorithms how to create neural network controllers for simulated car. They evolved controllers that have robust performance over different tracks and can be specialized to work better on particular tracks.

(Mahajan and Kaur, 2013) in "Neural Networks using Genetic Algorithms" introduced flexible method for solving the travelling salesman problem using genetic algorithms as they can be used to train neural networks producing evolutionary artificial neural networks.

(Fardin Ahmadizar et al, 2014) in "Artificial neural network development by means of a novel combination" developed a new evolutionary-based algorithm to simultaneously evolve the topology and the Connection weights of ANNs by means of a new combination of grammatical evolution (GE) and genetic algorithm (GA). GE is adopted to design the network topology while GA is incorporated for better weight adaptation. Please remember that all the papers must be in English and without orthographic errors.

## 3 SYSTEM OVERVIEW

A genetic algorithm (GA) (Vose, 1999) is an evolutionary algorithm that can solve optimization problems. It starts from a pool of randomly chosen candidate solutions of the optimization problem called a "population". Usually, a pre-knowledge about the problem constrains the randomness of these solutions. Each candidate solution is called a "chromosome". The algorithm repeatedly (over generations) modifies the population hoping for a new generation with a better population. For that, the algorithm uses an application-dependant "fitness function" that estimates the goodness of each chromosome. At each step, the genetic algorithm randomly selects individuals from the previous generation's population and uses them as parents to produce the children for the new generation. The concept of producing children from a set of selected parents is based on a natural selection process that mimics biological evolution. Hence, over successive generations, the population "evolves" toward an optimal solution.

Artificial neural networks can be looked at as an optimization problem looking for the best weights achieving some task. This is why a genetic algorithm can be used to train a neural network (Schaffer et al., 1992). Evolutionary Neural Networks, Neuroevolution, or neuro-evolution, is a form of machine learning that uses evolutionary algorithms to train artificial neural networks, in other words, estimating the weights of the neural network. It is most commonly applied in the areas of artificial life and intelligent computer games, and hence, has potential contributions towards self-driving vehicles. The chromosome format is chosen to be the vector of real numbers with a sequence of all of the neural network weights. The sequence is sorted layer by layer. The weights of each layer are sorted such that all of the weights coming out of a neuron are consecutive. The bias node is considered the last node in each layer. Figure 1 shows an example for a 2×3×2 neural network and its chromosome.

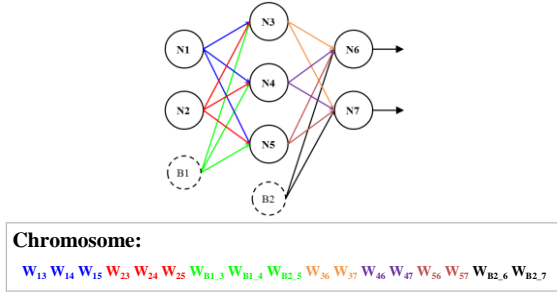

Figure 1: Example of a 2×3×2 neural network and its chromosome. B1 and B2 represent the network biases.

We developed a simulation setup to evaluate the fitness of each chromosome in a generation. For most of experiments related to collision avoidance, the vehicle lifetime before its first collision is a reasonable metric for the fitness. Genetic algorithm is used to pick and breed generations of more intelligent vehicles. The vehicle uses a rangefinder sensor that calculates $N$ intersections depths with the environment and then feeds these $N$ values as inputs to the neural network. The inputs are then passed through a multi-layered neural network and finally to an output layer of 2 neurons: a left and right steering force. These forces are used to turn the vehicle by deciding the vehicle steering angle. Figure 2 shows the proposed system overview for our method during the system training phase.

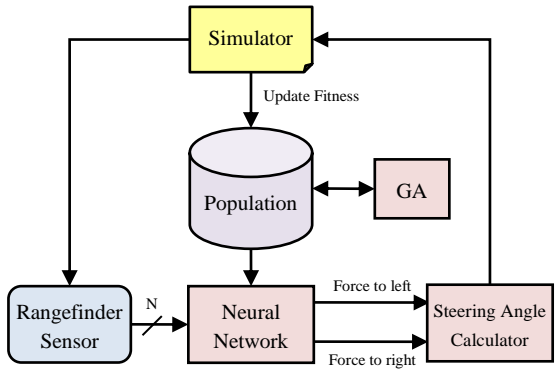

Figure 2: System overview during training phase.

Once trained, the neural network is able to generate steering commands from the input rangefinder sensor readings. Figure 3 shows this configuration.

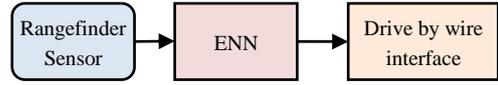

Figure 3: The trained network is used to generate steering commands from a single front-facing proximity sensor.

## 4 SIMULATION SETUP

For all vehicles in our simulation environment, we use a bicycle model as shown in figure 4. Given the vehicle speed and simulation time tick $\Delta t$, the travelled distance $L$ per a time step is calculated. Given wheel base, vehicle position $P$, heading $\theta$, and distance travelled per time step $L$, the new vehicle position $P_{new}$ and heading $\theta_{new}$ are calculated.

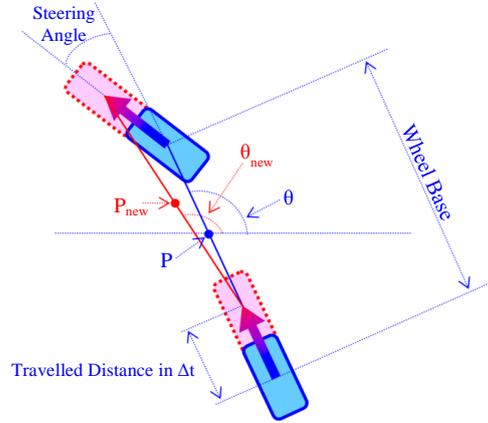

Figure 4: Simple 2D vehicle steering physics.

For simplicity, vehicles are chosen to move with fixed speed and sensor noise is neglected. At simulation start, the vehicles are positioned equidistant from each other. At each collision detected by the simulator, it's important to identify the vehicle responsible for the collision as shown in figure 5. When a collision happens, the simulator tries to answer the question: Would crash still happen if a vehicle $X$ is the only vehicle that moved at collision time step? If the answer is yes, vehicle $X$ is determined as responsible for that collision.

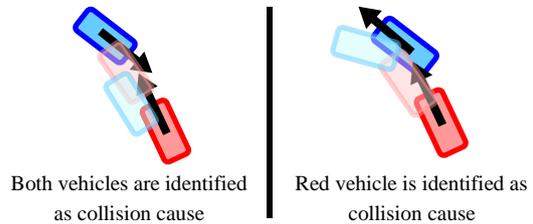

Both vehicles are identified as collision cause

Red vehicle is identified as collision cause

Figure 5: Determining which vehicle caused collision to happen. Two examples with two vehicles before and after the collision time step.

# 5 EXPERIMENTAL WORK

Several experiments are conducted to evaluate our learning method. The main objective is to inspect the feasibility of achieving a reactive collision avoidance system using our proposed method. Initially, an elementary, and relatively easier, experiment is conducted. The objective of this experiment is to examine the capability of a vehicle to learn the task of self-navigation through a static environment that does not include any dynamic objects. We believe that this task is less hard than the collision avoidance task because the ego-vehicle does not have to deal with the unknown movement of dynamic objects.

A three layer ANN, with sigmoid activations for all neurons, is empirically chosen to be used for all of our experiments. It's noted that the experiments results don't change if the number of layers is changed, but sometimes you obtain the same result faster. The higher the number of hidden layers, the better representation of the data the network can achieve. But at the same time, this leads to a more complex optimization problem that is harder and slower for GA to solve. Our GA uses a population of 200 chromosomes where mutation probability is 0.1, crossover probability is 1 and the crossover site follows a normal distribution with a mean of 0.95 and a standard deviation of 0.05. The selection is based on tournaments of size 10 candidates and children of next generations always replace their parents. The fitness function is chosen to be the vehicle lifetime navigating the environment (in time steps) before its first collision with the static environment boundaries or other dynamic vehicles.

The experimental work results are encouraging and validate the effectiveness of the proposed method.

## 5.1 Learning Navigation

The objective of this experiment is to validate the ability of our method to achieve self-navigation. The vehicle should learn how to travel from one position to another without colliding with a static environment that does not include any dynamic objects. The environment is represented by a track that is formed by horizontal and vertical edges. Figure 6 shows the experiment track.

Our experimental results show that navigation is learnt in less than 50 generations. One interesting observation is that the vehicle took 12 generations to learn how to successfully turn in the first critical location *A* circled in red in figure 6. Once the vehicle learns this manoeuvre, it achieves huge learning progress represented by a significant increase in best chromosome's fitness. The vehicle implicitly learns how to drive through all the following tricky turns in the track. This fact is demonstrated in figure 7.

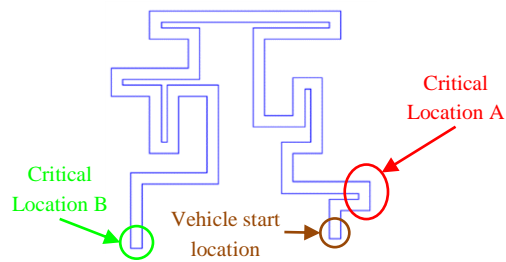

Figure 6: Experiment track.

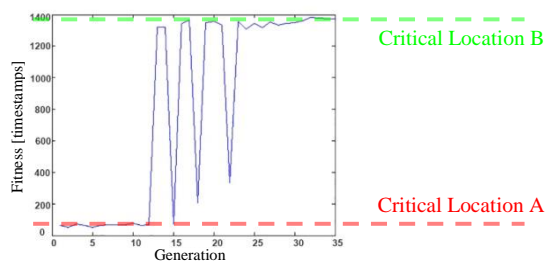

Figure 7: Self-navigation learning curve in a narrow track.

We can observe that the fitness diminishes when reaching critical location *B* circled in green in figure 6, the reason is that the vehicle modifies its behaviour in order to learn the 180° turn (location *B*) but what it learns negatively affect its ability to pass the previous critical location *A*, so the fitness oscillates until the vehicle learns to avoid such behaviour but it still unable to make the 180° turn. In that experiment, the track is too narrow, relative to the vehicle dimensions, which makes such move very hard to learn. Widening the track enables the vehicle learn how to turn by 180° and still be able to pass critical location *A* at the same time as demonstrated in figure 8.

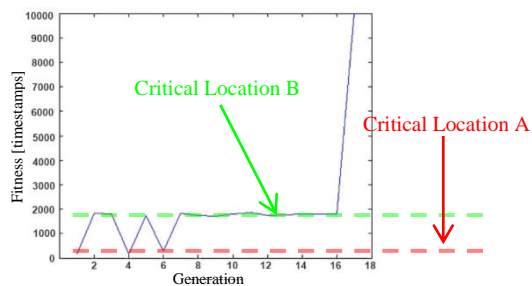

Figure 8: Self-navigation learning curve in a wide track.

As shown in figure 8, a very high fitness is reached as the vehicle learns to navigate back and forth through the track without any collisions for hours. The vehicle also is able to navigate successfully through other different tracks that are unseen during training.

## 5.2 Sensor Resolution

The objective of this experiment is to inspect the influence of the input rangefinder sensor resolution on the learning process. The same previous experiment is performed five times with different numbers number of sensor beams. The angle between adjacent beams is equal. The sensor horizontal range is chosen 180° in our experiment.

As shown in figure 9, using a sensor of a single front-facing beam prevents the vehicle from learning and reaching an accepted fitness as the input data is insufficient for learning. On the contrary, a higher sensor resolution (three beams or more) enables the vehicles to evolve and reach a satisfying fitness.

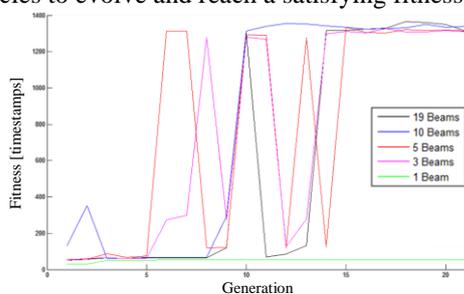

Figure 9: Self-navigation learning curve for different number of sensor beams.

It's observed that using a moderate number of beams achieved almost the same fitness as when using a larger number of beams but in a fewer number of generations. As in figure 9, the five beams' experiment is the fastest to reach an accepted fitness. This occurred because reducing the number of sensors produces shorter chromosomes and hence having a fewer number of parameters and a less complex optimization problem to solve.

## 5.3 Individual Collision Avoidance among Dynamic Vehicles

In this experiment, we inspect the feasibility of our method in enabling a vehicle to navigate collision-freely among multiple different dynamic vehicles. The environment is represented by a rectangular free space area containing eight different vehicles moving freely as shown in figure 10.

Each vehicle is initialized with random weights for their ANN and random starting headings then the learning process is applied on only a single vehicle to learn avoiding collisions with the environment boundaries and the other randomly moving vehicles.

We found that the learning vehicle (ego-vehicle) has learned a deceptive behaviour for survival by rotating around itself to avoid interactions with the other vehicles. In order to learn a proper collision avoidance behaviour, such rotation is detected by the simulator and the responsible chromosome is penalized by a zero fitness.

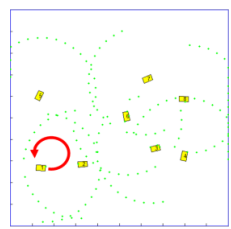

Figure 10: Ego-vehicle survives by rotating around itself.

After each collision occurs by the ego-vehicle, the fitness of the chromosome driving the neural network is estimated and a new chromosome is set to be evaluated. In order to achieve fair evaluation for each chromosome, not only the ego-vehicle should be reset but the whole simulation. Ignoring the reset of the simulation may position the ego-vehicle in tough scenario for collision avoidance at the beginning of evaluation. This may cause a good chromosome to be assigned a low fitness.

Each plot in figure 11 shows the learning curve of the ego-vehicle among uncontrolled dynamic vehicles. In each figure, a different movement strategy for the uncontrolled dynamic vehicles is adopted, and four runs of the same experiment are conducted. The x-axis represents the number of the generation and the y-axis represents the fitness achieved at each generation.

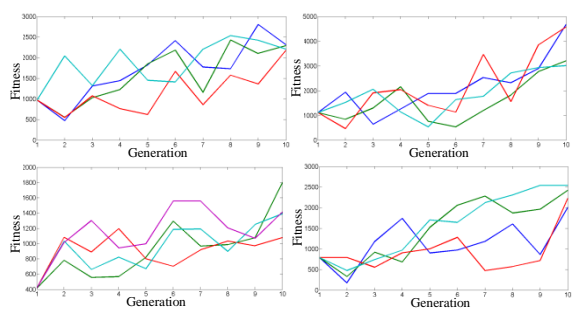

Figure 11: Each plot of the four represents a different strategy. In each strategy, the same experiment is conducted 4 times starting from different random initial neural network weights, each run is in a different color.

The learning curves demonstrate the ability of our method to enable the vehicle to learn collision avoidance individually among dynamic vehicles.

## 5.4 Individual Collision Avoidance Knowledge Accumulation

The objective of the experiment is to examine the knowledge accumulation ability of our method. In other words, the capability of the ego-vehicle to learn avoiding collisions in new strategies without negatively affecting the performance achieved in previously-learned strategies.

Firstly, as shown in Table 1, the ego-vehicle doesn't achieve efficient collision avoidance when tested on an unseen strategy. Learning on a single strategy is not sufficient for the ego-vehicle to learn general collision avoidance behaviour.

Table 1: Individual collision avoidance performance tested on unknown strategies compared to the performance tested on the strategy seen during training. The numbers in the table represents the best chromosome fitness.

| Training Strategy \ Deployment Strategy | 1 | 2 | 3 | 4 |
|---|---|---|---|---|
| 1 | 2115 | 544 | 558 | 432 |
| 2 | 595 | 2305 | 136 | 931 |
| 3 | 159 | 351 | 3050 | 460 |
| 4 | 559 | 334 | 1080 | 2560 |

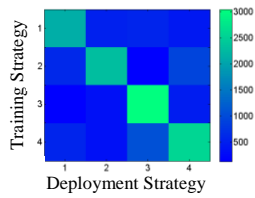

Figure 12: Color map visualization for fitness achieved in different strategies.

### 5.4.1 Incremental Evolution

As a step towards achieving general collision avoidance behaviour, the ego-vehicle should be trained on more than one strategy. Learning in a new strategy should not negatively affect the performance achieved in previously-learned strategies.

In order to achieve this objective, incremental evolution (Togelius and Lucas, 2006) is used. A vehicle learns to avoid collisions in one strategy and when it reaches an accepted fitness, a new strategy is added to the learning process so that the proposed solution is now evaluated on both strategies, then the fitness is averaged. This process is then repeated with new strategies added until the vehicle learns to survive in all introduced strategies. Table 2 shows the results of this experiment. The training stopping criteria for each strategy is when its fitness exceeds 80% of a predefined threshold for accepted fitness.

Table 2: Fitness achieved by the ego-vehicle during different incremental evolution iterations. A new strategy is added to the learning process at each iteration.

| Learning Iteration \ Deployment Strategies | 1 | 2 | 3 | Average Fitness |
|---|---|---|---|---|
| 1 | 1936 | - | - | 1936 |
| 2 | 2380 | 1452 | - | 1916 |
| 3 | 1970 | 2123 | 1467 | 1853.3 |

## 5.5 Simultaneous Collision Avoidance

The objective of this experiment is to achieve a collision free environment, where all moving vehicles simultaneously learn to avoid collisions with each other and with static environment. An evolved vehicle, that learned to navigate collision-freely, is used to boost the behaviour of the other vehicles through two different approaches as detailed in the coming two subsections. The results are obtained by running the simulator to train for 100 seconds in four different strategies.

### 5.5.1 Broadcasting the Winning Chromosome

In this approach, the evolved solution represented by the winning chromosome is broadcasted to all the vehicles to use. Table 3 compares the average number of collisions per second for all the vehicles before versus after learning.

Table 3: Comparison between the number of collisions per second before versus after learning.

|  | Initial Behaviour [collisions/sec] | After Learning [collisions/sec] |
|---|---|---|
| Strategy 1 | 16.33 | 3.90 |
| Strategy 2 | 20.05 | 4.59 |
| Strategy 3 | 16.82 | 8.00 |
| Strategy 4 | 13.49 | 12.27 |

In reasonable simulation time, the collision avoidance performance highly increases, but not for all the strategies.

### 5.5.2 Broadcasting the Most Evolved Generation

In order to achieve better collision avoidance performance, learning process should not only be applied on a single vehicle, but all vehicles should simultaneously learn. Instead of assigning the winning chromosome directly to each vehicle, we can assign the evolved population to each vehicle to start learning using it. This approach results in a customized solution for each different vehicle and our results are promising as the number of collisions is reduced by around 90% on the average.

Table 4: Comparison between the number of collisions per second before versus after learning.

|  | Initial Behaviour [collisions/sec] | After Learning [collisions/sec] |
|---|---|---|
| Strategy 1 | 16.33 | 2.11 |
| Strategy 2 | 20.05 | 1.58 |
| Strategy 3 | 16.82 | 1.76 |
| Strategy 4 | 13.49 | 1.90 |

## 5.6 Lane Keeping

The main objective of this experiment is to validate the generality of our method. A more realistic simulation environment is used. As shown in figure 13, the input is no longer readings from a proximity sensor, but lane markings from a camera. The objective is to achieve the lane keeping active safety feature given the detected driving lanes.

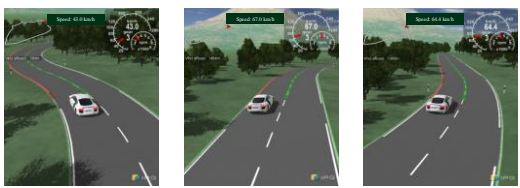

Figure 13: CarMaker simulation for lane keeping experiment

The results prove that our method generalizes well. The vehicle is left to learn on simulated roads for around 10 hours before it successfully learns to keep in a lane for many hours. It implicitly learnt many lane shape cases instead of memorizing a set of hardcoded scenarios. Our method successfully allows vehicle to learn different features other than collision avoidance like lane keeping, and using more realistic simulation environment.

## 6 CONCLUSIONS

This paper proposes and validates a novel method for vehicles reactive collision avoidance using ENN. To evaluate the proposed method, extensive experiments of varying conditions and objectives are conducted. The results demonstrated in the paper reflect the potential for our proposed method. The vehicle learns to drive collision freely in a static environment and among dynamic objects. Promising progress is achieved in developing general collision avoidance behaviour. Moreover, our lane keeping experiment shows the capability of our method to operate efficiently in realistic simulation environments. The future work should focus on deploying the conducted experiments in more realistic and complex simulation environments and to upgrade the GA operators to further improve our method's performance.